\relax
\documentclass[a4paper,11pt]{article}
\usepackage[square, sort]{natbib}
\usepackage[nottoc,notlot,notlof]{tocbibind}
\usepackage{times}  
\usepackage{helvet} 
\usepackage{courier}  
\usepackage{url}
\usepackage{fullpage}
\usepackage{graphicx}
\usepackage{graphics}
\usepackage{xcolor}
\definecolor{darkblue}{rgb}{0,0.08,0.8}

\usepackage{subcaption}
\usepackage{lmodern}
\usepackage{datatool}
\usepackage{algorithm}
\usepackage[noend]{algpseudocode}

\usepackage{amsthm,amsmath,amssymb}
\usepackage{bm}
\usepackage{latexsym}
\usepackage{float}
\usepackage{color}
\usepackage{pifont}
\newtheorem*{theorem*}{Theorem}

\newtheorem*{proposition*}{Proposition}

\newcommand{\A}{\mathcal{A}}
\newcommand{\bO}{\mathcal{O}}
\newcommand{\T}{\mathcal{T}}
\newcommand{\bS}{\mathcal{S}}
\newcommand{\E}{\mathbb{E}}
\newcommand{\D}{\mathcal{D}}
\usepackage[colorlinks=true,linkcolor=darkblue, citecolor=darkblue]{hyperref}

\usepackage{booktabs}       
\usepackage{nicefrac}       
\usepackage{microtype}      
\usepackage{multirow}
\usepackage{siunitx}

\usepackage{microtype}
\usepackage{mathtools}
\usepackage{adjustbox}
\usepackage{graphicx}
\usepackage[english]{babel}
\usepackage[utf8]{inputenc}
\usepackage{caption}

\newcolumntype{C}{>{$}c<{$}}
\newcolumntype{L}{>{$}l<{$}}

\begin{document}
\title{Batch-Augmented Multi-Agent Reinforcement Learning for Efficient Traffic Signal Optimization}
\author{Yueh-Hua Wu$^{1,2}$, I-Hau Yeh$^{3}$, David Hu$^{3}$, Hong-Yuan Mark Liao$^{2}$}
\date{
$^1$ National Taiwan University\\ 
\vspace{0.03in}
$^2$ Academia Sinica, Taiwan\\
\vspace{0.05in}
$^3$ Elan Corporation
}
\maketitle

\begin{abstract}
The goal of this work is to provide a viable solution based on reinforcement learning for traffic signal control problems. Although the state-of-the-art reinforcement learning approaches have yielded great success in a variety of domains, directly applying it to alleviate traffic congestion can be challenging, considering the requirement of high sample efficiency and how training data is gathered. 

In this work, we address several challenges that we encountered when we attempted to mitigate a serious traffic congestion occurring in a metropolitan area. Specifically, we are required to provide a solution that is able to (1) handle the traffic signal control when certain surveillance cameras that retrieve information for reinforcement learning are down, (2) learn from batch data without a traffic simulator, and (3) make control decisions without shared information across intersections.

We present a two-stage framework to deal with the above-mentioned situations. The framework can be decomposed into an Evolution Strategies approach that gives a fixed-time traffic signal control schedule and a multi-agent off-policy reinforcement learning that is capable of learning from batch data with the aid of three proposed components, bounded action, batch augmentation, and surrogate reward clipping.

We show that the evolution strategies method is able to obtain a fixed-time control schedule that outperforms reinforcement learning agents dynamically adjusting traffic light duration with much fewer samples from simulators.
For the multi-agent reinforcement learning part, the bounded action component maintains stability under the multi-agent scenario where controller at each intersection should make decision independently. 
The surrogate reward clipping enables multi-agent and off-policy reinforcement learning approaches to learn from batch data without a simulator in a cooperative task. 
Lastly, the batch augmentation mitigates the lack of efficiency in collecting data in traffic signal control problems.
Our experiments show that the proposed framework reduces traffic congestion by $36\%$ in terms of waiting time compared with the currently used fixed-time traffic signal plan. Furthermore, the framework requires only $600$ queries to a simulator to achieve the result. 
\end{abstract}

\section{Introduction}
Reinforcement learning (RL) as well as multi-agent reinforcement learning (MARL) are of great interest since plenty of real-world problems can be modeled as sequential decision-making problems. 
Although RL\footnote{For conciseness, RL is used to indicate both RL and MARL in the following content.} has showed its unprecedented capability of learning to achieve complicated tasks such as playing Go \citep{silver2017mastering}, it is, however, quite sample inefficient and usually requires a strong simulator so that the RL agent can learn from a large number of trial and error.
As a result, the real-world application of RL is still quite limited since a simulator that can well react as a real environment and is fast enough to generate sufficient data is usually unaffordable. 

Throughout the process of developing an efficient and portable RL solution that is compatible with traffic light controllers, we realize that some settings that previous work considered may be impractical due to its requirement of collecting real-time information from all intersections to make a thorough control decision and its authority of changing traffic signals at will. In practice, making control decision according to all observed information from intersections can be challenging since the transmission of local observations may be lagging or even shutdown. Besides, for better user experience, the countdown signals of traffic lights should be provided, which suggests that the number of the collected training data for RL learning would be much fewer and the decision made by the RL controllers should be able to handle traffic situations for more than a hundred seconds.

In this work, we address the following challenges that we encountered when applying RL and MARL approaches to alleviate traffic congestion in a metropolitan area: (1) collecting training data from simulators and real world can be slow especially when predicting the length of traffic signals is required; (2) RL methods are easily stuck at local optimum when solving traffic signal control for multiple adjacent intersections; and (3) learning in the on-policy or the off-policy setting is not always allowed since a traffic simulator and exploration for RL in real world are sometimes unavailable.

We propose a decentralized framework that allows traffic light controllers at different intersections to make decision independently. The framework is composed of an Evolution Strategies (ES) part and an MARL part. The ES method first optimizes a fixed-time traffic control schedule that serves as an initialization for the following MARL part. The advantages of using ES are twofold. The learning is much more sample efficient than directly applying RL methods and learning from scratch. Even though the result is merely a fixed-time schedule, it outperforms RL methods in terms of traffic control objective such as waiting time by a large margin. The other advantage is that the result is a fixed-time schedule that does not take real-time information as input; instead, it learns from the whole traffic flow distribution within a period, which provides a steady and reliable solution when surveillance cameras are down.

The MARL part can learn from a fixed-time control schedule in a batch RL fashion \citep{lange2012batch}; that is, the proposed MARL method does not need further interactions with simulators or the real world while training. The method is based on an off-policy MARL approach, MADDPG \citep{lowe2017multi}, that requires interactions with environments to update the estimate of its value function. To enable off-policy MARL approaches to learn from batch data, we start from discussing the current limitation of off-policy methods and attribute it to the overfitting to the given reward function. Then, we propose surrogate reward clipping that regularizes the learned value function and mitigates the overfitting problem. Besides, we propose batch augmentation that deals with the lack of training data by incorporating traffic knowledge and introducing additional information that RL methods can utilize. 

\subsection{Background}
In this section, we will first define the terms that are commonly used in traffic signal control literature. Then, we will introduce the problem setting of reinforcement learning and the corresponding objective that would be optimized to learn the desired tasks. Lastly, a brief introduction will be provided to bridge the gap between traffic signal control and reinforcement learning. The introduction is regrading what traffic information is usually adopted to serve as the input, output, and the objective in reinforcement learning.

\subsubsection{Term Definition in Traffic Signal Control}
To ensure the validity of the term definition adopted in this work, we refer to the definition given by \cite{wei2019survey}.
\begin{itemize}
    \item Movement signal: A movement signal is defined on the traffic movement, with green signal indicating the corresponding movement is allowed and red signal indicating the movement is prohibited.
    \item Phase: A phase is a combination of movement signals. To comprehend the concept of the phase in an intersection, one may produce a representation of the intersection by lining up all the traffic lights at that intersection with a certain order and such representation correspond to the phase. Different representations indicate different phases.
    \item Signal plan: A signal plan of an intersection is a sequence of phases $p_i$ with $t_i$ indicating the corresponding phase period. It can be represented by a set $\{(p_i,t_i)\}_{i=1}^n$ and $p_{i+1}$ occurs right after $p_i$. 
    \item Cycle-based signal plan: A cycle-based signal plan is a kind of signal plan where the sequence of phases operates in a cyclic order.
\end{itemize}

\subsubsection{Multi-Agent Reinforcement Learning and Markov Games}

Our setting is based on the standard Markov Decision Process (MDP) \citep{sutton1998introduction}. MDP is represented by a tuple $\langle \bS,\mathcal{A},\mathcal{P}, \mathcal{R},\gamma\rangle$, where $\bS$ is the state space, $\mathcal{A}$ is the action space, $\mathcal{P}(s_{t+1}\vert s_t,a_t)$ is the transition density of state $s_{t+1}$ at time step $t+1$ given action $a_t$ made under state $s_t$ at time step $t$, $\mathcal{R}(s,a)$ is the reward function, and $\gamma\in(0,1)$ is the discount factor.
An agent's behavior corresponds to a policy $\pi$ in reinforcement learning. A policy $\pi$ maps states to a probability distribution over the actions, $\pi : S\rightarrow \mathcal{P}(\mathcal{A})$. 
The performance of $\pi$ is evaluated in the $\gamma$-discounted infinite horizon setting and its expectation can be represented with respect to the trajectories generated by~$\pi$:
\begin{align}\label{eq:reward_sum}
    \mathbb{E}_\pi[\mathcal{R}(s,a)]=\mathbb{E}\left[\sum_{t=0}^\infty \gamma^t\mathcal{R}(s_t,a_t)\right],
\end{align}
where the expectation on the right-hand side is taken over the densities $p_0(s_0)$, $\mathcal{P}(s_{t+1}|s_t,a_t)$, and $\pi(a_t|s_t)$ for all time steps $t$.
Reinforcement learning algorithms~\citep{sutton1998introduction} aim to maximize Eq.~\eqref{eq:reward_sum} with respect to $\pi$.

In this work, we consider a multi-agent extension of MDPs called partially observable Markov games \citep{littman1994markov} since our goal is to develop a decentralized framework, where the information is not allowed to be shared among intersections. 
Similarly to MDP, the Markov game for $N$ agents is defined by a set of states $\bS_1, \bS_2,\cdots, \bS_N$ and a set of actions $\A_1, \A_2,\cdots,\A_N$. The states in this setting are only partially observable and what the agents are given is a set of observations $\bO_1,\bO_2,\cdots,\bO_N$. 
Each agent $i$, depending on the type of RL approaches, follows a stochastic policy $\pi_i:\bO_i\times\A_i\mapsto [0,1]$ or a deterministic policy $\mu_i:\bS\mapsto\A_i$ to choose an action. 
The next states will be produced according to the chosen actions, the states, and the state transition function $\T:\bS\times\A_1\times\cdots\times\A_N\mapsto  \bS$. 
Each agent $i$ aims to optimize its own total expected return $R_i=\sum_{t=0}^T\gamma^tr_i^t$, where $r_i:\bS\times\A_i\mapsto \mathbb{R}$ is a reward function, $\gamma$ is a discount factor, and $T$ is the time horizon.

\subsubsection{Multi-Agent Deep Deterministic Policy Gradient}\label{sec:maddpg}
\textbf{Policy Gradient.}\ \ \  Policy gradient algorithms are a popular choice for a variety of RL tasks and have been applied to multi-agent settings \citep{lowe2017multi}. Denote the state visitation density by policy $\pi$ as $\rho^\pi$.
The main idea is to update the policy $\pi$ parameterized by $\theta$ so that the objective $J(\theta)=\mathbb{E}_{s_\sim\rho^\pi,a\sim\pi}\left[R\right]$ is maximized, by taking gradient steps in the direction of $\nabla_\theta J(\theta)$:
\begin{align}\label{eq:pg}
    \nabla_\theta J(\theta)=\E_{s\sim\rho^\pi,a\sim\pi}\left[\nabla_\theta\log\pi(a\vert s)Q^\pi(s,a)\right],
\end{align}
where $Q^\pi$ is the value function that is used to estimate the total expected return of following policy $\pi$. Several practical algorithms have been proposed by adopting different approaches to estimate $Q^\pi$ \citep{williams1992simple, sutton2018reinforcement}.

\noindent \textbf{Deterministic Policy Gradient (DPG).}\ \ \  It is of interest for policy gradient (Eq.~\ref{eq:pg}) to consider the case when the policy is deterministic~\citep{silver2014deterministic}. With a slight abuse of notation, we denote that the deterministic policy $\mu$ is parameterized by $\theta$. We may rewrite Eq.~\ref{eq:pg} under the condition that the action space  $\A$ is continuous:
\begin{align}\label{eq:dpg}
    \nabla_\theta J(\theta)=\E_{s\sim \D}\left[\nabla_\theta\mu(a\vert s)\nabla_aQ^\mu(s,a)\vert_{a=\mu(s)}\right],
\end{align}
where $\mathcal{D}$ is the replay buffer where DPG stores state samples with previous policies.

\noindent \textbf{Deep Deterministic Policy Gradient (DDPG).}\ \ \  DDPG is based on the above deterministic policy gradient. DDPG is an off-policy approach that utilizes neural networks as function approximators for the policy $\mu$ and the value function $Q^\mu$.
To stabilize the learning of the value function, the use of the target network is proposed in DDPG. Since the update of the value function is via minimizing a loss in a recursive fashion,
\begin{align}\label{eq:ddpg}
    \mathcal{L}=\E_{s_t\sim\rho^\beta,a_t\sim\beta}\left[(Q^\mu(s_t,a_t)-y_t)^2\right],
\end{align}
where $y_t=r(s_t,a_t)+\gamma Q^\mu(s_{t+1},\mu(s_{t+1}))$ and $\beta$ is a stochastic policy for exploration, the change of the target $y_t$ keeps the value function from converging to a proper local optima. The proposed target network replaces the value function used to estimate $y_t$ with a delayed value function so that the target value will not be altered too much during training.

\noindent \textbf{Multi-Agent Deep Deterministic Policy Gradient (MADDPG).}\ \ \ 
There are different multi-agent settings related to information sharing for MARL. Some approaches consider a completely independent case where the agents are totally isolated at the training and the inference phases. 
MADDPG is operated upon the setting where the information sharing is allowed in the training phase but the agents can only observe local information at execution time. Therefore, similarly to \cite{foerster2018counterfactual}, MADDPG is a framework of centralized training with decentralized execution. 

The motivation behind MADDPG is the non-stationary environment induced from the other agents that prevents traditional single-agent policy gradient from converging to an optimal policy. 
Consider a game with $N$ agents with policies parameterized by $\pmb{\theta}=\{\theta_1,\cdots,\theta_N\}$ and let $\pi=\{\pi_1,\cdots,\pi_N\}$. For agent $i$, since the other agents can be regarded as a part of the environment, the fact that each agent is changing during training results in optimization under a dynamic environment.
From another aspect, if the actions taken by all agents are known, the environment can still remain stationary as the policies change since $\Pr(s'\vert s,a_1,\cdots,a_N,\pi_1,\cdots,\pi_N)=\Pr(s'\vert s,a_1\cdots,a_N)=\Pr(s'\vert s,a_1,\cdots,a_N,\pi_1',\cdots,\pi_N')$ for any $\pi_i\neq \pi_i'$.
Based on the above observation, MADDPG proposes to extend the value function to a centralized state-value function $Q_i=(x,a_1,\cdots,a_N)$, where $x=(o_1,\cdots,o_N)$ consist of the observations of all agents.
The policy gradient of the expected return for agent $i$ can then be expressed as,
\begin{align}
    \nabla_{\theta_i}J(\theta_i)=\E_{s\sim p^\mu,a_i\sim\pi_i}\left[\nabla_{\theta_i}\log \pi_i(a_i\vert o_i)Q_i(x,a_1,\cdots,a_N)\right].
\end{align}

Since MADDPG is straightforward and easy to implement, we use MADDPG as the underlying MARL method in our experiments. However, it should be noted that the proposed Batch Augmentation and Surrogate Reward Clipping are compatible with any off-policy RL approaches and also MARL in cooperative scenarios.

\subsection{Reinforcement Learning for Traffic Signal Control}
To utilize RL to optimize traffic signal control, it is required to determine the objective of the control problem and discover the corresponding reward function. In addition to the reward function, the features that potentially give a correct description of a traffic network are also discussed in some previous work. 

There are some possible reward functions such as queue length, waiting time, and change of delay and the choice of the reward function determines the task the RL agents attempt to learn. To take care of every aspect, we may also optimize a weighted linear combination of all the components.
The state features are also an important factor for RL agents to learn well-behaved policies. Without informative states, it would be harder for RL agents to make an inference from the received observation $o$ to the real state $s$. There are some common options to represent traffic situations such as queue length, volume, speed, and position of vehicles.

In this work, we propose a state- and reward-agnostic framework, which means that we do not focus on a certain type of state or reward and it is capable of handling a variety of kinds of states and rewards such as queue length and change of delay. With the aid of neural network architectures, it is also straightforward to handle high-dimensional states such as multiple images by adding convolution layers to the function approximators for the policy and value networks.

We study a specific type of action space, cycle-based signal plan, for traffic signal control. There are some common action definitions \citep{wei2019survey} in previous work such as:
\begin{itemize}
    \item Keep or change the current phase in a fixed order of a signal plan.
    \item Keep or choose the next phase to change.
    \item Set current phase duration.
    \item Set cycle-based phase ratio.
    \item Set cycle-based phase length. The difference in determining the length and the ratio is that setting the ratio indicates that the total length is fixed and setting the length has more flexibility of choosing a better cycle length.
\end{itemize}
In this work, setting cycle-based phase length for multiple intersections at the beginning of a cycle is the action. Although keeping or changing the current seems to be a straightforward action space for RL, this kind of action fails to provide countdown signals for drivers and results in worse user experience empirically. Furthermore, since the currently used fixed-time signal controllers take the signal length as input, to make the developed algorithm compatible with the controllers and to prevent additional cost at designing new controllers, setting cycle-based phase length is the best option among the common action spaces. 
Setting cycle-based length, however, is relatively difficult to optimize since it takes a longer time to collect a single state-action pair. For instance, when the action is to keep or to change the current phase, to collect one state-action pair, it takes merely a sampling step that is usually between 1 to 10 seconds. On the other hand, the cycle length can be up to 100 to 200 seconds at a major avenue, and gathering the state-action pairs become rather time-consuming.

To sum up, from the standpoint of providing easy-to-use and economic algorithm, we develop a framework that is able to optimize the cycle-based phase length and to learn from much fewer state-action pairs compared with the standard MARL method.

\subsection{Problem Setting}\label{sec:probdef}
Denote the given phase sequences for $N$ intersections as $\{\{p_{i,j}\}_{i=1}^{n_j}\}_{j=1}^N$, where $n_j$ is the number of phases of intersection $j$. The action is the corresponding phase length $\pmb{a}_t=\{\{t_{i,j}\}_{i=1}^{n_j}\}_{j=1}^N$. 
To make the action space more manageable and to synchronize intersections for better traffic flow, we impose a length constraint such that
\begin{align}
    \sum_{i=1}^{n_j}t_{i,j}=\sum_{i=1}^{n_z}t_{i,z},&&\text{ for }j\neq z.
\end{align}

The objective we aim to optimize is the total reward sum $\E_{s\sim \rho_\pi,a\sim\pi}[R]$ as mentioned in Section~\ref{sec:maddpg}. It can be any factors such as queue length or change of delay. In our experiments, we maximize the negative average waiting time,
\begin{align}
    R=&\sum_{t=0}^T r(s_t,a_t)\\
    =&-\sum_{t=0}^T \E_{s_t\sim\rho_\pi,a_t\sim\pi}\left[\frac{1}{\ell(a_t) } \text{WaitingTime}(s_t,a_t)\right]\label{eq:obj},
\end{align}
where 
\begin{align}
    \ell(a_t)=\frac{\text{length of $a_t$ in seconds}}{\text{sampling length $\delta_s$ in seconds}},
\end{align}
and $\pi=\{\pi_1,\pi_2,\cdots,\pi_N\}$ is a set of all policies at each intersection. The policies are decentralized and sharing information at execution time is not allowed. It should be noted that the symbol $t$ in Eq.~\ref{eq:obj} indicates the number of cycles. The WaitingTime function gives the summation of waiting time of vehicles in the targeted traffic network during $a_t$.

\section{Related work}\label{relatedwork}
To the best of our knowledge, this work presents the first framework that optimizes the cycle-based signal plan in the decentralized multi-agent scenario. Furthermore, the proposed MARL method is able to directly learn from batch data, which suggests its potential to improve the current traffic signal control plan without simulators.

Some previous work considers a similar action space but for only one intersection. To reduce action space, \cite{liang2018deep} proposes to estimate the length difference between the next cycle and the current one but it considers that only one phase is changed and the amount of the difference is fixed. For example, if there are three phases and the corresponding phase length are $(t_1, t_2, t_3)$, then the possible length combinations for the next cycle are $(t_1\pm\triangle t, t_2\pm\triangle t, t_3\pm\triangle t)$ and $(t_1,t_2,t_3)$ with $\triangle t$ some predefined value. The action space in \cite{casas2017deep} is the cycle-based phase ratio and the cycle length is decided via grid search. Although \cite{casas2017deep} aims to solve traffic congestion for multiple intersections, it assumes that the controllers are centralized and the proposed algorithm can be realized by a single agent.



\section{Proposed Method}
\begin{figure}
    \centering
    \includegraphics[scale=0.5]{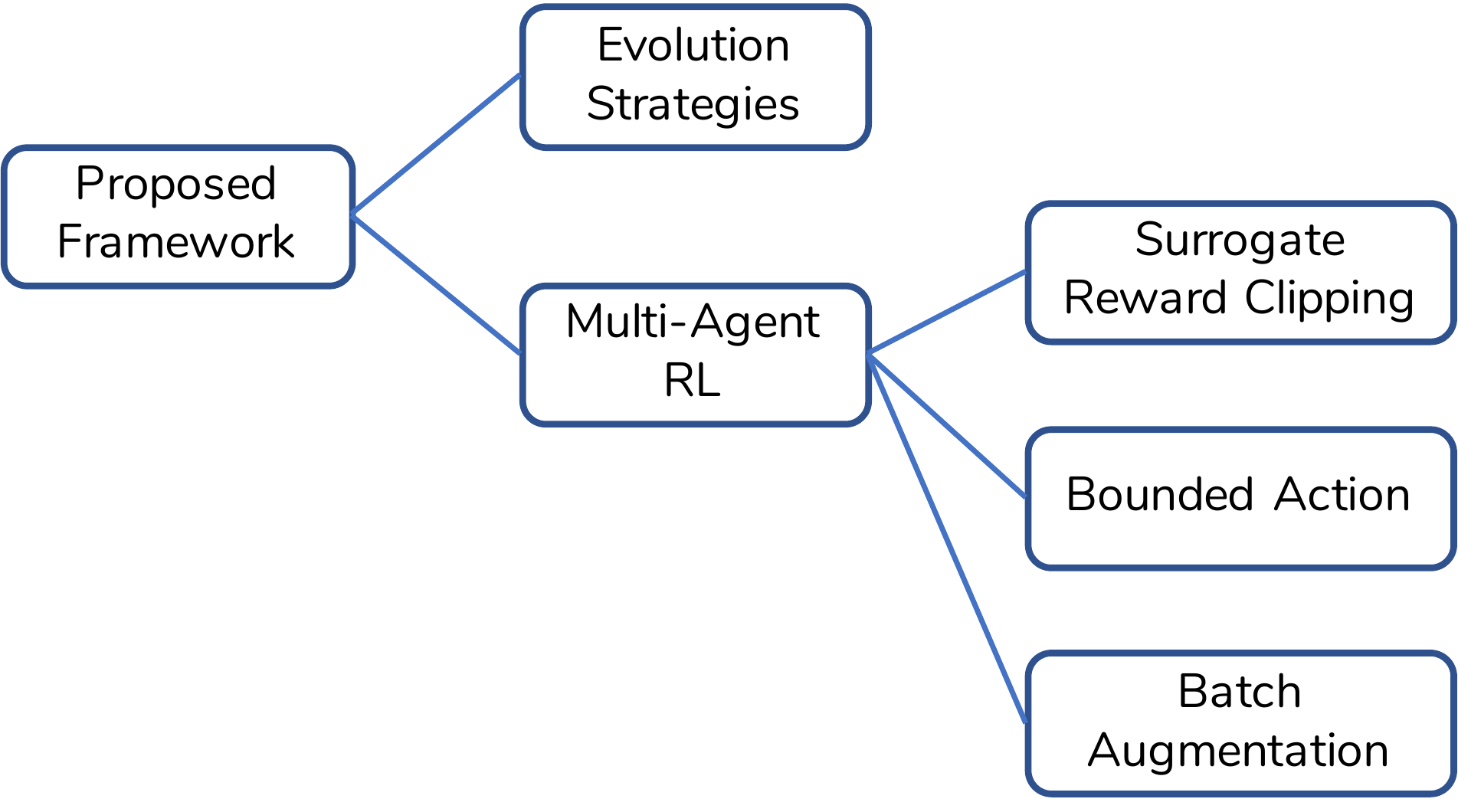}
    \caption{Framework details.}
    \label{fig:structure}
\end{figure}

In this section, we will present our framework as shown in Figure.~\ref{fig:structure}. The proposed Evolution Strategies for fixed-time traffic signal optimization (ES) provides a stable and sample-efficient algorithm to optimize the length of each phase for multiple intersections. Note that the total cycle length is changeable and we empirically found that the cycle length plays an important role in traffic signal control. 

The proposed MARL method is able to learn an adaptive control system from a fixed-time control schedule. To enhance its generalizability and applicability, we consider the batch RL setting \citep{lange2012batch}, where interacting with simulators or the real world is not allowed. This setting is especially suitable for traffic signal control since creating a simulator that is able to perfectly imitate the real transition in the real world is quite costly and the simulation rate may be too slow to satisfy the sample complexity of typical RL methods. Toward this end, we propose three components, surrogate reward clipping, bounded action, and batch augmentation, which empower the approach's potential to learn from batch data in our experiments with much fewer data than the MADDPG baseline. 
Concretely, the MADDPG is not able to attain a multi-agent policy that is able to surpass the original fixed-time control plan with the same amount of data and the proposed components obtain a multi-agent policy that reduces waiting time by $16\%$ in our experiments.

Since we use neural networks as the function approximators for the policy and the value function, thanks to a variety of neural network architectures, the proposed method is capable of handling a wide range of state and reward representations. The states can be given in the form of high-dimensional sequential representations such as images and temporal data, respectively. To learn from such data, we may simply incorporate convolutional layers \citep{krizhevsky2012imagenet,kim2014convolutional,karpathy2014large} or transformer \citep{jaderberg2015spatial} to our function approximators. In other words, the proposed framework including the ES and the MARL parts is both state-agnostic and reward-agnostic, which suggests that the users of our framework are free to utilize any task-specific information and network architecture.

\subsection{Evolution strategies for fixed-time traffic signal optimization}
Evolution strategies (ES) \citep{rechenberg1994evolutionsstrategie} is a particular set of black-box optimization algorithms, which is indifference to the distribution of the fitness function it attempts to optimize. Specifically, the ES algorithms we use in this work belong to natural evolution strategies (NES) \citep{glasmachers2010natural,glasmachers2010exponential,schaul2011high}. NES considers a population with a distribution over parameters $p_\phi(\theta)$ and the goal of NES is to maximize the average fitness function over the distribution $\E_{\theta\sim p_\phi}\left[F(\theta)\right]$. The parameter $\theta$ usually represents weights of a neural network in optimization problems. 

To utilize NES to solve traffic signal optimization, we propose to directly concatenate all phase length to represent the signal state of the current traffic network: $\theta=(t_{1,1},t_{2,1},\cdots,t_{n_1,1},\cdots,t_{n_N,N})$. The fitness function, as mentioned in Section~\ref{sec:probdef}, is average negative waiting time.

To learn to maximize $F(\theta)$, we may parameterize the distribution $p_\phi$ as an isotropic multivariate Gaussian with mean $\phi$ and fixed covariance $\sigma^2I$, which gives another form of the average fitness function:
\begin{align}
    \E_{\theta\sim p_\phi}[F(\theta)]=\E_{\epsilon\sim N(0,I)}[F(\theta+\sigma\epsilon)].
\end{align}
To optimize the objective, we may derive the gradient with the aid of the score function estimator \citep{salimans2017evolution}:
\begin{align}\label{eq:ES}
    \nabla_\theta \E_{\epsilon\sim N(0, I)}[F(\theta+\sigma\epsilon)]=\frac{1}{\sigma}\E_{\epsilon\sim N(0,I)}[F(\theta+\sigma\epsilon)\epsilon].
\end{align}
Since the gradient in Eq.~\ref{eq:ES} can be empirically estimated by gathering samples and calculating (simulating) the fitness value, the NES algorithm optimizes the fitness function $F(\theta)$ by altering between two phases \citep{salimans2017evolution}: (1) sampling perturbations $\epsilon_1,\epsilon_2,\cdots,\epsilon_n\sim \mathcal{N}(0, I)$ and (2) updating parameters, 
\begin{align}\label{eq:ES_update}
    \theta_{t+1}\leftarrow \theta_t+\alpha\frac{1}{n\sigma}\sum_{i=1}^nF_i\epsilon_i,
\end{align}
where $\alpha$ is learning rate and $n$ is the number of samples.

To employ NES to optimize the length of the phases at each intersection, it should be noted that there are a number of constraints for the parameter $\theta$:
\begin{enumerate}
    \item to synchronize each intersection for better traffic flow, the cycle length at each intersection are the same;
    \item the length of each phase is lower and upper bounded for reasonable traffic signal design.
\end{enumerate}
As a result, we should put more effort in sampling perturbations for traffic signal phases. We propose a sampling method exclusively for optimizing fixed-time traffic signal control:
\begin{enumerate}
    \item find the intersection $i$ with the least number of phases,
    \item sample perturbations for each phase $\triangle t_{1,i},\cdots,\triangle t_{n_i,i}$ from $\mathcal{N}(0,\sigma)$ at intersection $i$ and clip the values so that $t_{1,i}+\triangle t_{1,i},\cdots,t_{n_i,i}+\triangle t_{n_i,i}$ are within the valid range,
    \item compute the sum of the perturbations $\triangle t=\sum_{j=1}^{n_i}\triangle t_{j,i}$,
    \item for each intersection $k\neq i$, sample perturbations for phases $1$ to $n_k-1$ from $\mathcal{N}(0,\sigma)$ and the perturbation for the last phase is determined by $\triangle t-\sum_{z=1}^{n_k-1}\triangle t_{z,k}$; clipping the phase difference so that the new phases are valid.
\end{enumerate}
The reason why we aim to choose the intersection with the least number of phases to compute the difference sum is because the variance of the sum $n_i\sigma^2$ is in proportion to the number of phases. Therefore, if the chosen $n_i$ is too much larger than $n_j$, it is equivalent to using large learning rate and the update would be less stable.

In the above-mentioned perturbation method, all perturbations are sampled from a normal distribution $\mathcal{N}(0,\sigma)$ and since we maintain the same cycle length for all intersections, the scale of the perturbation of the last phase may be different. To handle the mismatch of perturbation scale within an intersection, an alternative way is to apply intersection-conditioned variance. Instead of sampling from $\mathcal{N}(0,\sigma)$, for each intersection $i$, we may obtain the perturbation from $\mathcal{N}(0,\frac{\sigma}{\sqrt{n_i}})$ so that the variance of the sum of the perturbations at each intersection is approximately $\sigma^2$. The method is summarized below:
\begin{enumerate}
    \item randomly choose an intersection $i$,
    \item sample perturbations for each phase $\triangle t_{1,i},\cdots,\triangle t_{n_i,i}\sim\mathcal{N}(0, \frac{\sigma}{\sqrt{n_i}})$ at intersection $i$ and clip the values so that $t_{1,i}+\triangle t_{1,i},\cdots,t_{n_i,i}+\triangle t_{n_i,i}$ are within the valid range,
    \item compute the sum of the perturbations $\triangle t=\sum_{j=1}^{n_i}\triangle t_{j,i}$,
    \item for each intersection $k\neq i$, sample perturbations for phases $1$ to $n_k-1$ from $\mathcal{N}(0,\frac{\sigma}{n_k})$ and the perturbation for the last phase is determined by $\triangle t-\sum_{z=1}^{n_k-1}\triangle t_{z,k}$; clipping the phase difference so that the new phases are valid.
\end{enumerate}

Although the proposed perturbation is no longer i.i.d. and Eq.~\ref{eq:ES} may not hold, the real gradient can be approximated by Eq.~\ref{eq:ES} when the variance $\sigma^2$ is smaller. Empirically, we will show in the experiments that such approximation is accurate enough to consistently improve the performance.

\subsection{Multi-Agent Reinforcement Learning from Batch Data}

In this section, we present an MARL method that is capable of learning from batch data. Specifically, we propose three components that enable off-policy MARL approaches such as MADDPG to learn from batch data without suffering from extrapolation error \citep{fujimoto2018off}.

To be more precise, the problem setting of our MARL approach is that given a fixed-time signal control plan $\pi_\mathrm{fix}$ and state-action pairs sampled with $\pi_\mathrm{fix}$, our goal is to find an adaptive control policy that is decentralized at execution time and outperforms $\pi_\mathrm{fix}$. The setting is practical and allows us to easily improve the currently used traffic signal plan without additional construction of an accurate simulator or effort in collecting new data. 

Although it may sound trivial that adaptive policies are superior to the fixed ones, empirically there are some challenges induced from empirical traffic concerns that hinder us from directly employing state-of-the-art MARL approaches and attaining better results.
Examples are the above-mentioned decentralized decision making and the difficulty in gathering a large amount of training data due to the time-consuming action space.

\subsubsection{bounded action}
To learn from a set of state-action pairs, a straightforward way is to utilize behavior cloning~\citep{schaal1999imitation} (BC) or batch RL~\citep{lange2012batch} (BRL) to derive a similar or superior policy to the one collecting those pairs. However, we found that to imitate the given policy, directly adopting BC and BRL methods and learning from scratch can easily fall into local optimums that are much worse than the given policy. We attribute this observation to the fact that a reasonable plan for traffic networks should properly arrange traffic lights so that traffic flows can proceed from one intersection to another. Although the concept is simple, it is hard for RL controllers to pick up the idea mostly because the viable solution space gets smaller when the traffic network becomes larger; as we consider more intersections, the influence of phase conflict among intersections will be amplified. This explains why the currently employed fixed-time signal plan is likely to perform better than RL results.

As a result, we propose to predict a bounded phase difference instead of the whole phase length. Concretely, the original MARL produce phase length $\pmb{a}_t=\{\{t_{i,j}\}_{i=1}^{n_j}\}_{j=1}^N$ at the beginning of each phase cycle to maximize the return and we propose to produce the \emph{phase length difference} $\pmb{a}'_t=\{\{\triangle t_{i,j}\}_{i=1}^{n_j}\}_{j=1}^N$ and the phase length can be recovered via 
$$\pmb{a}_t=\{\{b_{i,j}+\triangle t_{i,j}\}_{i=1}^{n_j}\}_{j=1}^N,$$
where $\{\{b_{i,j}\}_{i=1}^{n_j}\}_{j=1}^N$ is a fixed-time traffic plan. We may directly use the currently deployed traffic signal plan or the result from the ES method as $\{\{b_{i,j}\}_{i=1}^{n_j}\}_{j=1}^N$. The value of $\triangle t_{i,j}$ is bounded by $\pm\delta$ and $\delta>0$.

There are a number of advantages of learning the phase difference. The first aspect is that to alleviate traffic congestion, the traffic signal should be reliable and usually predictable. Learning to give a bounded difference also bound the phase difference between the last cycle and the current one (at most $2\delta$). Another advantage is that it enables the decentralized traffic system to become more trustworthy. Since the traffic information is not shareable while the adaptive system is deployed, it is possible that the other agents do not respond as expected, which may result in serious blockage. The use of the bounded difference greatly reduces such risk to a tolerable level. Lastly, we will show in the proposed batch augmentation that the bound of the phase difference $\delta$ influences the quality of the approximation and provides a trade-off between output flexibility and learning efficiency.

\subsubsection{batch augmentation}

To correctly represent situations of a traffic network, various kind of elements have been proposed to describe the environment such as queue length, waiting time, volume, etc. As a result, the state representation is usually of high dimension. As discussed in previous sections, the neural network architectures of the policies are capable of handling high dimensional state space but the estimation of reward function still rely on standard sampling on high dimensional space. Concretely, since the action of our problem setting is the cycle-based phase length, it takes several sampling steps to compute the $\text{WaitingTime}$ function in Eq.~\ref{eq:obj},
\begin{align}\label{eq:smalltime}
    \text{WaitingTime}(s_t,a_t) = \sum_{i=0}^{\ell(a_t)-1}\text{WaitingTime}(s_{t+\frac{i}{\ell(a_t)}\delta_s},a'_{t+\frac{i}{\ell(a_t)}\delta_s}),
\end{align}
where $a'_{t+\frac{i}{\ell(a_t)}\delta_s}$ is an action that indicates the current phase that lasts $\lvert a'_{t+\frac{i}{\ell(a_t)}\delta_s}\rvert= \delta_s$.
This result is derived from the fact that the MDP ($\mathcal{M}$) we consider consist of a smaller MDP ($\mathcal{M}'$), where decision making occurs at every sampling step. 

Although it is valid to utilize sampling to acquire $\text{WaitingTime}(s_t,a_t)$, the stochastic environment and high dimensional state representation result in high estimation variance. The empirically high variance is one of the reasons for sample inefficiency. An example is that if we already know that distribution has low or no variance, then usually few samples are required to draw the outline of such distribution and vice versa. As a result, we propose a data augmentation method that incorporates traffic knowledge and enables us to learn sample enhanced value function for multiple phases. The additional value functions substantially reduce the variance of policy gradient estimates at the cost of some bias. The discussion about the trade-off between estimation bias and variance for RL can be found in \cite{schulman2015high}.

In the following content, we will first introduce the batch augmentation method for intersections with two phases and will further extend the method to the cases with multiple phases.

\textbf{Two Phases.} \ \ \ 
Before introducing batch augmentation, we would like to reiterate that the update of the value function of off-policy RL depends on the exploration policy $\beta$ instead of the agent policy $\mu$ as indicated in the expected value of Eq.~\ref{eq:ddpg}. In other words, the update of the value function can be comprehended as a way to improve understanding of the consequences and the knowledge regarding the environment. As a result, although it takes much time to collect a single state-action pair when the action is to set the cycle-based phase length, we may utilize the consequences given in the state-action pair to provide more knowledge to the value function.

\begin{figure}
    \centering
    \includegraphics[scale=0.8]{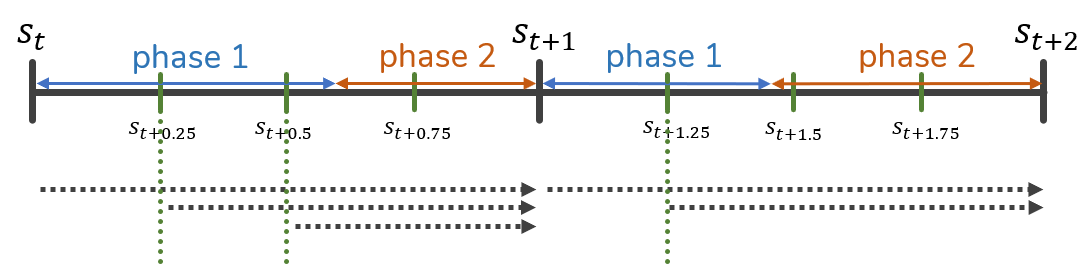}
    \caption{Batch augmentation for two phases.}
    \label{fig:ba2p}
\end{figure}

An example is illustrated in Figure~\ref{fig:ba2p}. 
For better clarity, we will use a state and the next state to represent a state-action pair.
Suppose that the intersection has only two phases and the sampling step length equals $0.25$ cycle length.
Originally, in the given example, we have only two state-action pairs, $(s_t, s_{t+1})$ and $(s_{t+1}, s_{t+2})$. In addition to the two pairs, we may also infer a consequence $(s_{t+0.25}, s_{t+1})$, which means that if we reduce the length of phase 1 with $0.25$ cycle length and begin from $s_{t+0.25}$, we will end up with $s_{t+1}$. Similarly, we may enhance the given data and the number of pairs increases from $2$ to $5$. It should be noted that we cannot directly infer $(s_{t+0.75},s_{t+1})$ or $(s_{t+1.5},s_{t+2})$ since those consequences do not include information about phase 1. Consequently, the above-mentioned batch augmentation may work well for intersections that have few phases but the influence becomes limited as the number of phases increases. To address the problem, we propose to learn \emph{truncated value functions} in the following discussion for multiple phases.

\begin{figure}
    \centering
    \includegraphics{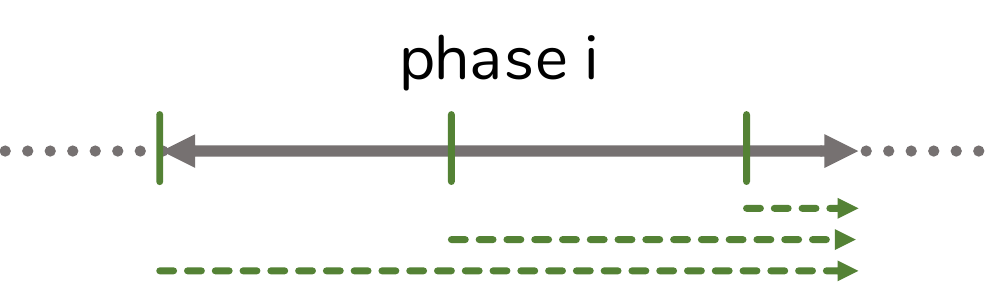}
    \caption{The green dotted lines indicate augmented data that can be utilized to train the truncated value function $Q^\mu_i$.}
    \label{fig:multiphases}
\end{figure}

\textbf{Multiple Phases.} \ \ \ To utilize the batch augmentation method for intersections with multiple phases, we may first decompose the original value function with respect to different phases. Since this method can cope with common reward function used for traffic signal control such as queue length, speed, and waiting time, for better representability, we replace the negative \text{WaitingTime} function in Eq.~\ref{eq:obj} and Eq.~\ref{eq:smalltime} with a reward function $f_r(s,a)$. Let $n$ denote the number of phases, the action $a_t=(t_1,\cdots,t_n)$ represents the length of each phase, and $s'_i\coloneqq s_{t+\frac{\sum_{j=0}^{i-1}t_j}{\sum_{j=1}^nt_j}}$. 
We may rewrite the expected value of the value function $Q^\mu$:
\begin{align}
    Q^\mu(s_t,a_t)=&\E_{s_{t+1}\sim \D}\left[r(s_t,a_t)+\gamma Q^\mu(s_{t+1},\mu(s_{t+1}))\right]\\
    =&\E_{s_{t+1}\sim \D}\left[\frac{1}{\ell(a_t)}f_r(s_t,a_t)+\gamma Q^\mu(s_{t+1},\mu(s_{t+1}))\right]\\
    =&\E_{s_{t+1}\sim \D}\left[\frac{1}{\ell(a_t)}\sum_{i=1}^{n}f_r\left(s'_i, t_{i}\right)+\gamma Q^\mu(s_{t+1},\mu(s_{t+1}))\right]\\
    =&\E_{s_{t+1}\sim \D}\left[\frac{1}{\ell(a_t)}\sum_{i=1}^n  t_i\Bar{Q}^\mu_i+\gamma Q^\mu(s_{t+1},\mu(s_{t+1}))\right]&&\left(\Bar{Q}^\mu_i\coloneqq\frac{1}{t_i}f_r(s'_i,t_i)\right)\\
    =&\E_{s_{t+1}\sim \D}\left[\frac{1}{\ell(a_t)}\sum_{i=1}^n  (b_i+\triangle t_i)\Bar{Q}^\mu_i+\gamma Q^\mu(s_{t+1},\mu(s_{t+1}))\right]\\
    \stackrel{b_i\gg \triangle t_i}{\approx}& \E_{s_{t+1}\sim \D}\left[\sum_{i=1}^n  \left(\frac{b_i}{\ell(a_t)}\right)\Bar{Q}^\mu_i+\gamma Q^\mu(s_{t+1},\mu(s_{t+1}))\right]\label{eq:ba},
\end{align}
where $\Bar{Q}_i^\mu$ is called the \emph{truncated value function} and it is essentially the average of rewards collected within phase $i$. It should be noted that since we consider the multi-agent setting, the state $s_i'$ is composed of observations from different intersection. The advantages of the use of the truncated value function are twofold. Incorporating the truncated value function to estimate the average reward enables us to reduce the estimation variance, which improves the sample complexity of the underlying MARL approach. The other advantage is that the truncated value function can utilize batch augmentation to gather more data as shown in Figure~\ref{fig:multiphases}. For each simulation step, we can collect the corresponding state, action, and average reward as a training pair, which is not limited to the first phase. To implement the method, it is not necessary to train $n-1$ networks for the truncated value functions; instead, we found that the learned model would be more generalized if we train only one model which takes state $s$, action $a$, and a phase code $c$ as inputs. The phase code $c$ indicates which phase the model is referring to.

To sum up, the proposed batch augmentation is able to apply to intersections with multiple phases. Specifically, we utilize the first batch augmentation method to enhance the replay memory $\D$ to obtain $\D'$ and the expected value in Eq.~\ref{eq:ba} can be estimated with respect to $\D'$:
\begin{align}
    Q^\mu(s_t,a_t) \approx \E_{s_{t+1}\sim \D'}\left[\sum_{i=1}^n  \left(\frac{b_i}{\ell(a_t)}\right)\Bar{Q}^\mu_i+\gamma Q^\mu(s_{t+1},\mu(s_{t+1}))\right],
\end{align}
where the truncated value function $\Bar{Q}_i^\mu$ can be learned by adopting the second batch augmentation method to acquire more accurate estimates.

\subsubsection{off-policy learning from batch data via surrogate reward clipping}\label{sec:SRC}
Off-policy RL approaches claim to be capable of updating the agent policy $\mu$ with another exploration policy $\beta$ since the expected value of the value function is independent of the agent policy. Although the approaches seem to be promising to the batch RL setting, directly employing off-policy RL approaches to learn from data gathered with some unknown policy does not obtain decent performance as expected.

\cite{fujimoto2018off} claims that previous off-policy approaches are not truly off-policy due to the extrapolation error, a phenomenon in which unseen state-action pairs are erroneously estimated to have unrealistic values. When the state-action pairs are collected from a policy that is much different from the agent policy, the mismatch in the distribution of data makes the empirical estimate of the expected value inaccurate and harmful to the update of the agent policy.

To prevent the neural network estimator from estimating via extrapolation, we proposed Surrogate Reward Clipping (SRC) to constrain the learning of the value function. 
We observe that off-policy RL approaches are able to learn from the batch data from the beginning of the training iterations, which suggests that off-policy approaches are still able to capture knowledge about the tasks to some extent. As the learning proceeds, off-policy approaches may become unstable and the performance can even degrade, which is aligned with the observation of \cite{fujimoto2018off}. We approach this phenomenon from the perspective of overfitting.
In typical RL settings, an RL agent is able to update its belief to the world by collecting interactions with the environment. On the other hand, in the batch RL setting, gathering new state-action pairs is not allowed and overfitting may occur when an RL agent keeps optimizing its objective, the reward function.
Based on the observation, we develop SRC for cooperative multi-agent scenarios, which leverages reward shaping to regularize RL optimization without the need to carefully tune the weight.

In this method, we consider the reward function that can be used to evaluate any types of traffic networks, which is common in the domain of traffic signal control such as waiting time, queue length, throughput, etc.
Concretely, to estimate the performance of a method that attempts to alleviate traffic congestion, average waiting time at the traffic network can be a viable option no matter what the network size is, which suggests that average negative waiting time is a proper reward function for our method.

Since a traffic network is a decomposable structure that consists of multiple smaller networks, the use of the average negative waiting time can as well be applied to the networks at the lower level of the hierarchy. Although there are a number of possibilities of such hierarchies, to apply SRC, we may consider a two-level hierarchy where each intersection represents a sub-network in the lower level. The reward function for intersection $i$ at time $t$ is denoted as $r_{i,t}$ and the one for the whole network $r_{g,t}$. It should be noted that directly optimizing $r_{i,t}$ for each intersection will not lead to optimal policies.

The motivation behind SRC is to regularize the learning of the value function with $r_i$ and a weight $\alpha$,
\begin{align}
    R'_i=\sum_{t=0}^T\gamma^t(\alpha r_{g,t} + (1-\alpha) r_{i,t}),
\end{align}
where $0<\alpha<1$.
However, it may be tricky and inefficient to determine the optimal value of $\alpha$. Overly small $\alpha$ results in agents ignoring the importance of cooperation and too large $\alpha$ makes the additional term useless. In addition, the optimal value of $\alpha$ may alter according to different traffic networks and such value is hard to be directly obtained.
Therefore, we propose to clip $r_{i,t}$ with an upper bound so that the system will tend to improve all the agents,
\begin{align}\label{eq:src}
    R''_i=\sum_{t=0}^T\gamma^t(\alpha r_{g,t} + (1-\alpha) \min(r_{i,t},c_i)).
\end{align}

Another advantage of SRC is that it is easier to determine the value of $c_i$ than the weight $\alpha$ because the physical meaning of $c_i$ is clear -- it filters out a proportion of the surrogate rewards $r_{i,t}$ so that the new reward function $R''_i$ is less sensitive to the extreme values of the surrogate rewards. Consequently, we may directly set the value of $c_i$ according to statistics such as the second or the third quartile. 

Given Eq.~\ref{eq:src}, we may rewrite the value function:
\begin{align}
    Q^\mu =& \E_{\rho^\mu}\left[R''_i\right]\\
    =&\E_{s_t\sim\rho^\mu}\left[\sum_{i=0}^T\gamma^t(\alpha r_{g}(s_t, \mu(s_t))+(1-\alpha) \min(r_{i}(s_t,\mu(s_t)),c_i))\right]\\
    =&\E_{s_t\sim\rho^\mu}\left[\sum_{i=0}^T\gamma^t\alpha r_{g}(s_t, \mu(s_t))\right]+\E_{s_t\sim\rho^\mu}\left[\sum_{i=0}^T\gamma^t(1-\alpha) \min(r_{i}(s_t, \mu(s_t)),c_i)\right]\\
    =&\alpha\E_{s_t\sim\rho^\mu}\left[\sum_{i=0}^T\gamma^t r_{g}(s_t, \mu(s_t))\right]+(1-\alpha)\E_{s_t\sim\rho^\mu}\left[\sum_{i=0}^T\gamma^t \min(r_{i}(s_t, \mu(s_t)), c_i)\right]\\
    =&\alpha\left(Q^\mu_g+\frac{(1-\alpha)}{\alpha}\underbrace{Q^\mu_i}_{\text{regularization term}}\right),
\end{align}
where $Q_g^\mu\coloneqq\E_{s_t\sim\rho^\mu}\left[\sum_{i=0}^T\gamma^t r_{g}(s_t, \mu(s_t))\right]$ and 
\begin{align}
    Q_i^\mu\coloneqq\E_{s_t\sim\rho^\mu}\left[\sum_{i=0}^T\gamma^t \min(r_{i}(s_t, \mu(s_t)), c_i)\right]\leq c_i\frac{1-\gamma^{T+1}}{1-\gamma}.
\end{align}
We may interpret the result from the perspective of regularization. The value function $Q_i^\mu$ is an upper bounded regularization term with weight $\frac{1-\alpha}{\alpha}$. The regularization term prevents the policy from overfitting to the given batch dataset and obtains much better results in our experiments. Although it is possible to fine tune the weight $\frac{1-\alpha}{\alpha}$ when a simulator is accessible, it should be noted that it is a batch MARL setting and accessibility to a simulator is prohibited. 
Therefore, we are not able to evaluate the performance for multiple times to choose the best set of parameters. Instead of relying on parameter search, the proposed upper-bounded regularization empirically show a stable learning curve without additional tuning of parameters.

\section{Experiments}
In this section, we will present the experiment results of the proposed framework. To validate the effectiveness of each proposed component, the experiments are arranged into two parts, evolution strategies and ablation studies for MARL components. In the ES experiments, since the proposed method is capable of optimizing both cycle length and phase length, we will present the learning curve of the ES method as well as visualize the relationship among learning performance, phase length, and cycle length.
For the proposed MARL components, the experiment is an ablation study that reveals how each element improve the learning of MADDPG from batch data in terms of waiting time.

In our experiments, we consider a traffic network in a metropolitan area suffering from traffic congestion. The network consists of five intersections and the numbers of phases are $(n_1,n_2,n_3,n_4,n_5)=(5,3,2,2,2)$. The time interval we consider in our experiments is between five-thirty and six o'clock in the afternoon when traffic congestion is the most serious due to the afternoon rush hour. 
To optimize the fixed-time traffic control plan and to collect state-action pairs for the next MARL training, we adopt VISSIM~\citep{fellendorf2010microscopic} as our traffic flow simulator.

\subsection{Evolution Strategies}
To improve learning efficiency and reduce variance, we utilize antithetic sampling \citep{geweke1988antithetic} that samples pairs of perturbations $\epsilon$ and $-\epsilon$. 
Due to the length constraint of each phase, the corresponding counterpart of the perturbation may not necessarily exist. For example, if the current phase length is $12$ seconds and the sampled perturbation $\epsilon=4$, we are unable to directly apply $-\epsilon$ to derive a new phase since phase length is lower bounded by $10$ seconds. For such a case, we will use $16$ and $10$ seconds as the new phase length. We sample $10$ pairs from the above perturbation to form a new generation.

The scale of fitness function also requires careful handling. As shown in Eq.~\ref{eq:ES_update}, the gradient term $\alpha\frac{1}{n\sigma}\sum_{i=1}^n F_i\epsilon_i$ is directly related to the scale of $F_i$ but it may be different from that of $\theta$. Although it is possible to adjust the scale via choosing a proper learning rate $\alpha$, the process may be tricky and time-consuming. 
To resolve the scale mismatch, we found that the fitness shaping that applies a rank transformation~\citep{wierstra2014natural} to the fitness function before each parameter update is particularly useful.
\begin{figure}
    \centering
    \includegraphics[scale=0.30]{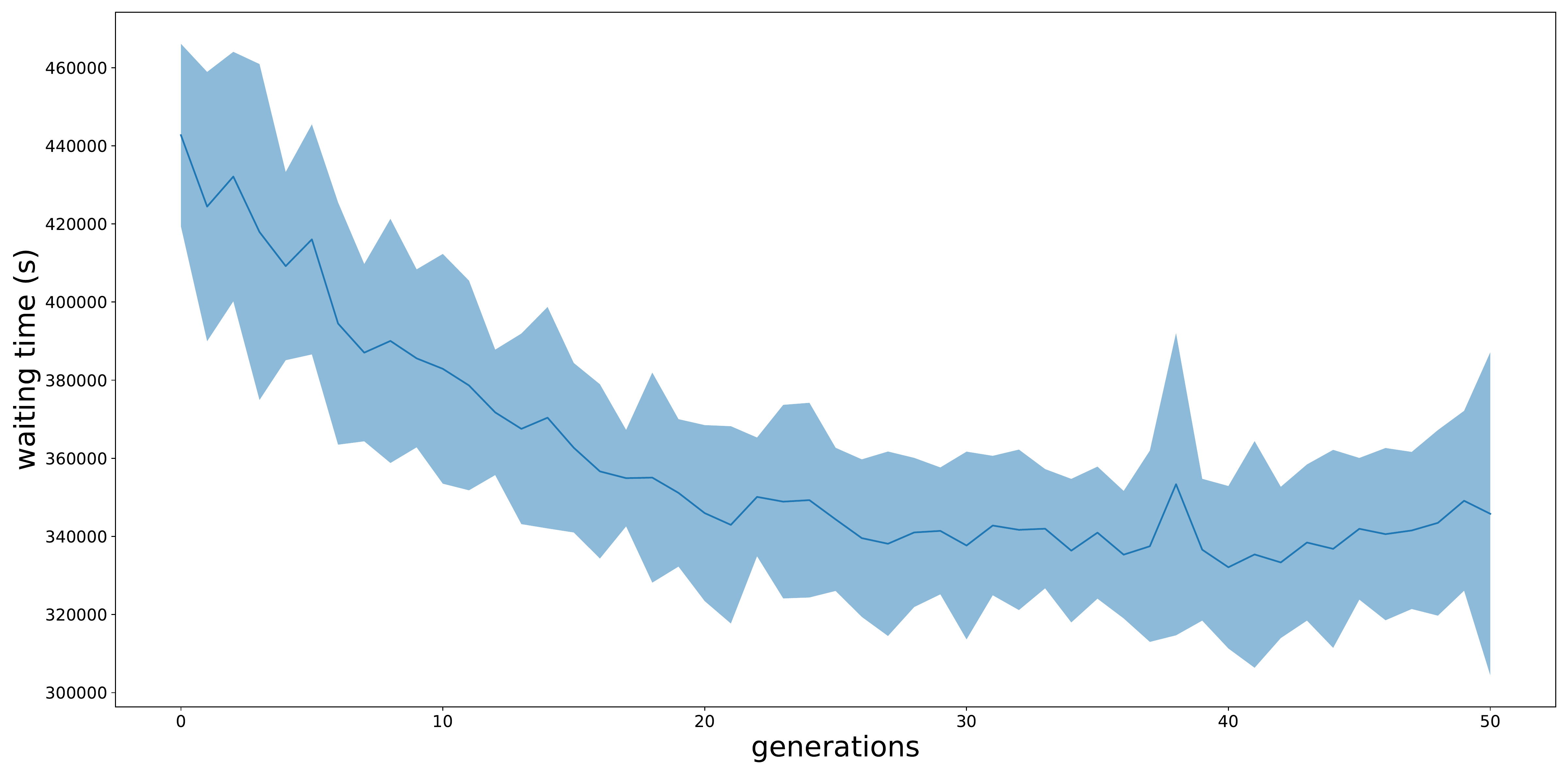}
    \caption{Learning curve of the proposed Evolution Strategies. The lower waiting time the better.}
    \label{fig:learningcurve}
\end{figure}

We set the initialization of the parameters $\theta$ to be the currently used fixed-time traffic signal control plan so that the difference can be easily observed. Since we directly adopt phase length as the parameters $\theta$, the optimization process is sample-efficient as shown in Figure~\ref{fig:learningcurve}. The ES method reduces the average waiting time by around $25\%$ in $30$ generations, which is equivalent to merely $600$ queries the fitness function.

\begin{figure}
    \centering
    \includegraphics[scale=0.3]{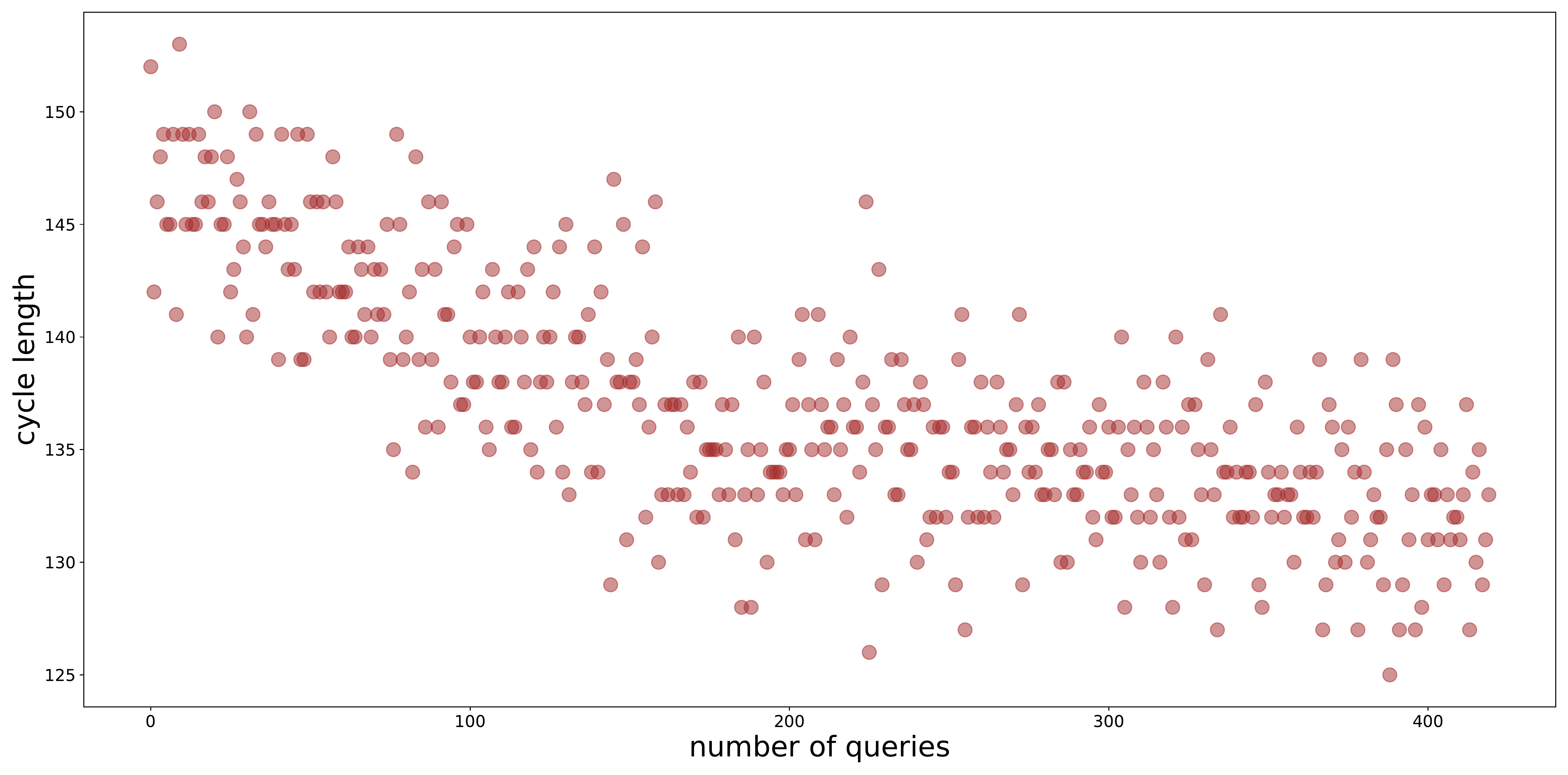}
    \caption{Learning curve with respect to cycle length.}
    \label{fig:exploration}
\end{figure}

\begin{figure}
    \centering
    \includegraphics[scale=0.3]{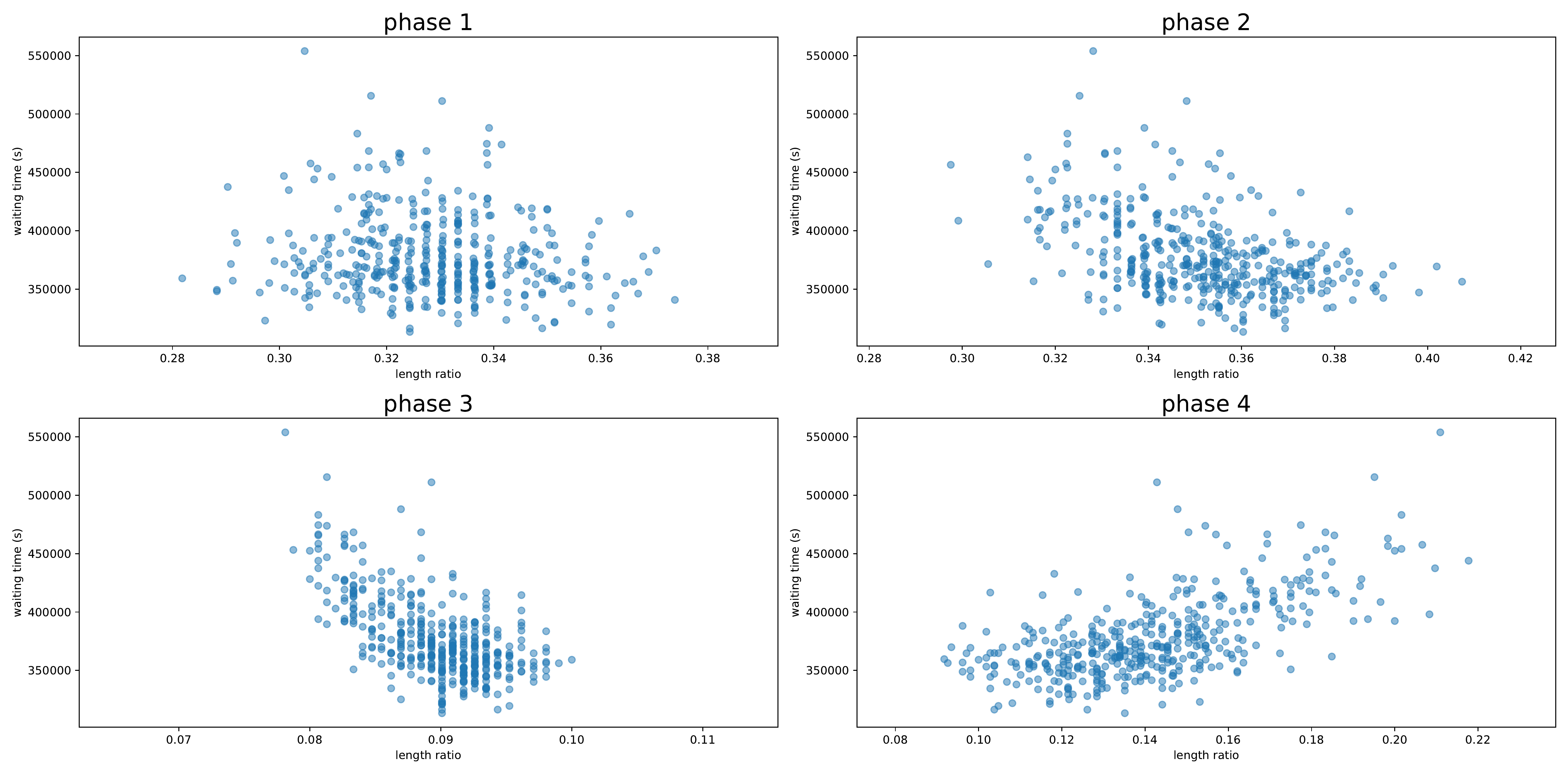}
    \caption{The distributions of the phase ratio and the corresponding waiting time for the first four phases at intersection $1$ observed during learning. The lower waiting time the better.}
    \label{fig:phaseratiol}
\end{figure}

The distributions of cycle length and each phase length is visualized in Figure~\ref{fig:exploration} and Figure~\ref{fig:phaseratiol}. At the beginning of learning, the ES method discovers that the cycle length is too long and the proposed perturbation enables it to explore for more suitable length for the traffic network. In addition to cycle length, the ratio of each phase is of importance to improve traffic flows. We observe from the distribution plots in Figure~\ref{fig:phaseratiol} that enlarging phase ratio of phase $2$ and phase $3$ reduces average waiting time effectively, which is aligned with the fact that the second and the third phases allow huge traffic flow to pass through. Although the ES method has no background knowledge regarding the traffic structure and the phase information, it is still able to find a proper distribution for each phase length as well as cycle length, through the process of pursuing lower waiting time.

\subsection{An Ablation Study for Multi-Agent Reinforcement Learning Components}

\begin{figure}
    \centering
    \includegraphics[scale=0.4]{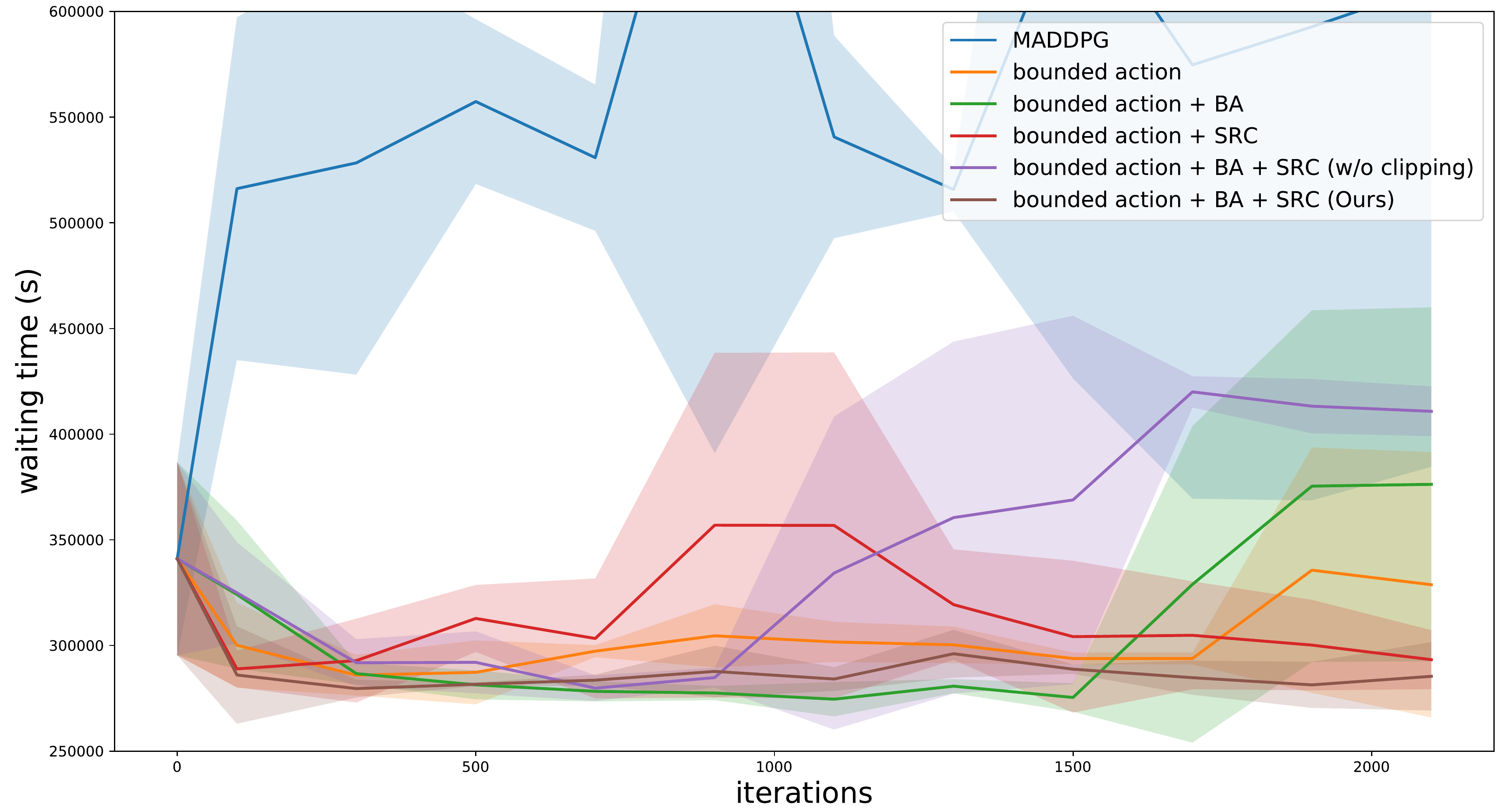}
    \caption{Learning curves of the effectiveness of each proposed component. For conciseness, we omit the MADDPG in the legend for the other five curves. The lower waiting time the better.}
    \label{fig:reward_comp}
\end{figure}

In Figure~\ref{fig:reward_comp}, we show a throughout ablation study of each proposed component. We use MADDPG as the underlying off-policy MARL method and the learning curve shows that the baseline is unable to learn an adaptive system superior to its initialization. As discussed above, since the system is decentralized at execution, the decision making can easily collapse when the action space is not well constrained. 

The ablation study shows that the influence of batch augmentation and SRC is clearly different. We may first compare the methods with and without batch augmentation. The methods with batch augmentation such as the green and the brown line, are able to derive shorter waiting time compared with their counterpart. Since the use of batch augmentation provides more traffic consequences regarding the traffic network, those incorporating batch augmentation are able to better optimize the given objective.
On the other hand, SRC improves MARL from the perspective of preventing overfitting. Even though methods with batch augmentation can achieve lower waiting time, the curves soon become upward-sloping and obtain worse performance.
As in Section~\ref{sec:SRC}, we approach the problem from the aspect of \emph{reward overfitting} since in the batch setting, interacting with an environment is not allowed and the belief learned from the given batch data cannot be corrected. 
Therefore, the methods with SRC, the red and the brown curves, are regularized with the surrogate reward acquired from local networks, which shows stability toward reducing waiting time without going off the rails. 

Lastly, the method equipped with the three proposed components is able to decrease waiting time by $15\%$ without suffering from instability during training even in the batch setting. It should be noted that the waiting time is initialized by the result of the ES method. Therefore, compared with the currently used traffic control plan, the whole framework reduces waiting time by $35\%$ with merely $600$ queries to a simulator.

\section{Conclusion}
In this work, we have proposed a sample-efficient framework for traffic signal control optimization. Specifically, our goal is to provide a solution that is able to provide a fixed-time control plan to handle situations where surveillance cameras are down, to learn a cooperative control system without communication between controllers at each intersection, and to achieve the above-mentioned goals with as few samples as possible.  

The framework is composed of two parts, an evolution strategies method, and a multi-agent reinforcement learning method. 
In the evolution strategies method, we have proposed a perturbation approaches sufficing the constraint that the cycle length should remain the same for all intersection. Besides, the proposed perturbation handles variance inconsistency for each phase within an intersection and attain better stability.
The evolution strategies method optimizes the phase length and the cycle length to obtain a fixed-time traffic control plan that achieves $25\%$ improvement compared with the currently used fixed-time traffic control plan in terms of waiting time.

For the multi-agent reinforcement learning, we address several empirical problems, the instability under the multi-agent scenario that communication is prohibited, the reward overfitting in the batch reinforcement learning setting, and the sample inefficiency due to the time-consuming action space. We have proposed three components, bounded action, surrogate reward clipping, and batch augmentation, respectively, to handle the above-mentioned problems. 
The off-policy multi-agent algorithm can improve a fixed-time traffic plan and learn from batch data without additional interactions with a simulator or the real world. The method is not only sample-efficient since no interactions with an environment are required but also data-efficient from the perspective of the number of the batch data.
We have shown in the experiment section with a thorough ablation study regarding the effectiveness of each proposed components and the method equipped with all the components are capable of further improving the result of the evolution strategies by $15\%$, drawing $36\%$ improvement in total compared with the currently used fixed-time traffic plan.


\bibliography{reference}

\begin{thebibliography}{27}
\providecommand{\natexlab}[1]{#1}
\providecommand{\url}[1]{\texttt{#1}}
\expandafter\ifx\csname urlstyle\endcsname\relax
  \providecommand{\doi}[1]{doi: #1}\else
  \providecommand{\doi}{doi: \begingroup \urlstyle{rm}\Url}\fi

\bibitem[Casas(2017)]{casas2017deep}
Noe Casas.
\newblock Deep deterministic policy gradient for urban traffic light control.
\newblock \emph{arXiv preprint arXiv:1703.09035}, 2017.

\bibitem[Fellendorf and Vortisch(2010)]{fellendorf2010microscopic}
Martin Fellendorf and Peter Vortisch.
\newblock Microscopic traffic flow simulator vissim.
\newblock In \emph{Fundamentals of traffic simulation}, pages 63--93. Springer,
  2010.

\bibitem[Foerster et~al.(2018)Foerster, Farquhar, Afouras, Nardelli, and
  Whiteson]{foerster2018counterfactual}
Jakob~N Foerster, Gregory Farquhar, Triantafyllos Afouras, Nantas Nardelli, and
  Shimon Whiteson.
\newblock Counterfactual multi-agent policy gradients.
\newblock In \emph{Thirty-second AAAI conference on artificial intelligence},
  2018.

\bibitem[Fujimoto et~al.(2018)Fujimoto, Meger, and Precup]{fujimoto2018off}
Scott Fujimoto, David Meger, and Doina Precup.
\newblock Off-policy deep reinforcement learning without exploration.
\newblock \emph{arXiv preprint arXiv:1812.02900}, 2018.

\bibitem[Geweke(1988)]{geweke1988antithetic}
John Geweke.
\newblock Antithetic acceleration of monte carlo integration in bayesian
  inference.
\newblock \emph{Journal of Econometrics}, 38\penalty0 (1-2):\penalty0 73--89,
  1988.

\bibitem[Glasmachers et~al.(2010{\natexlab{a}})Glasmachers, Schaul, and
  Schmidhuber]{glasmachers2010natural}
Tobias Glasmachers, Tom Schaul, and J{\"u}rgen Schmidhuber.
\newblock A natural evolution strategy for multi-objective optimization.
\newblock In \emph{International Conference on Parallel Problem Solving from
  Nature}, pages 627--636. Springer, 2010{\natexlab{a}}.

\bibitem[Glasmachers et~al.(2010{\natexlab{b}})Glasmachers, Schaul, Yi,
  Wierstra, and Schmidhuber]{glasmachers2010exponential}
Tobias Glasmachers, Tom Schaul, Sun Yi, Daan Wierstra, and J{\"u}rgen
  Schmidhuber.
\newblock Exponential natural evolution strategies.
\newblock In \emph{Proceedings of the 12th annual conference on Genetic and
  evolutionary computation}, pages 393--400, 2010{\natexlab{b}}.

\bibitem[Jaderberg et~al.(2015)Jaderberg, Simonyan, Zisserman,
  et~al.]{jaderberg2015spatial}
Max Jaderberg, Karen Simonyan, Andrew Zisserman, et~al.
\newblock Spatial transformer networks.
\newblock In \emph{Advances in neural information processing systems}, pages
  2017--2025, 2015.

\bibitem[Karpathy et~al.(2014)Karpathy, Toderici, Shetty, Leung, Sukthankar,
  and Fei-Fei]{karpathy2014large}
Andrej Karpathy, George Toderici, Sanketh Shetty, Thomas Leung, Rahul
  Sukthankar, and Li~Fei-Fei.
\newblock Large-scale video classification with convolutional neural networks.
\newblock In \emph{Proceedings of the IEEE conference on Computer Vision and
  Pattern Recognition}, pages 1725--1732, 2014.

\bibitem[Kim(2014)]{kim2014convolutional}
Yoon Kim.
\newblock Convolutional neural networks for sentence classification.
\newblock \emph{arXiv preprint arXiv:1408.5882}, 2014.

\bibitem[Krizhevsky et~al.(2012)Krizhevsky, Sutskever, and
  Hinton]{krizhevsky2012imagenet}
Alex Krizhevsky, Ilya Sutskever, and Geoffrey~E Hinton.
\newblock Imagenet classification with deep convolutional neural networks.
\newblock In \emph{Advances in neural information processing systems}, pages
  1097--1105, 2012.

\bibitem[Lange et~al.(2012)Lange, Gabel, and Riedmiller]{lange2012batch}
Sascha Lange, Thomas Gabel, and Martin Riedmiller.
\newblock Batch reinforcement learning.
\newblock In \emph{Reinforcement learning}, pages 45--73. Springer, 2012.

\bibitem[Liang et~al.(2018)Liang, Du, Wang, and Han]{liang2018deep}
Xiaoyuan Liang, Xunsheng Du, Guiling Wang, and Zhu Han.
\newblock Deep reinforcement learning for traffic light control in vehicular
  networks.
\newblock \emph{arXiv preprint arXiv:1803.11115}, 2018.

\bibitem[Littman(1994)]{littman1994markov}
Michael~L Littman.
\newblock Markov games as a framework for multi-agent reinforcement learning.
\newblock In \emph{Machine learning proceedings 1994}, pages 157--163.
  Elsevier, 1994.

\bibitem[Lowe et~al.(2017)Lowe, Wu, Tamar, Harb, Abbeel, and
  Mordatch]{lowe2017multi}
Ryan Lowe, Yi~I Wu, Aviv Tamar, Jean Harb, OpenAI~Pieter Abbeel, and Igor
  Mordatch.
\newblock Multi-agent actor-critic for mixed cooperative-competitive
  environments.
\newblock In \emph{Advances in neural information processing systems}, pages
  6379--6390, 2017.

\bibitem[Rechenberg(1994)]{rechenberg1994evolutionsstrategie}
Ingo Rechenberg.
\newblock Evolutionsstrategie: Optimierung technischer systeme nach prinzipien
  der biologischen evolution. frommann-holzbog, stuttgart, 1973.
\newblock \emph{Step-Size Adaptation Based on Non-Local Use of Selection
  Information. In Parallel Problem Solving from Nature (PPSN3)}, 1994.

\bibitem[Salimans et~al.(2017)Salimans, Ho, Chen, Sidor, and
  Sutskever]{salimans2017evolution}
Tim Salimans, Jonathan Ho, Xi~Chen, Szymon Sidor, and Ilya Sutskever.
\newblock Evolution strategies as a scalable alternative to reinforcement
  learning.
\newblock \emph{arXiv preprint arXiv:1703.03864}, 2017.

\bibitem[Schaal(1999)]{schaal1999imitation}
Stefan Schaal.
\newblock Is imitation learning the route to humanoid robots?
\newblock \emph{Trends in Cognitive Sciences}, 3\penalty0 (6):\penalty0
  233--242, 1999.

\bibitem[Schaul et~al.(2011)Schaul, Glasmachers, and
  Schmidhuber]{schaul2011high}
Tom Schaul, Tobias Glasmachers, and J{\"u}rgen Schmidhuber.
\newblock High dimensions and heavy tails for natural evolution strategies.
\newblock In \emph{Proceedings of the 13th annual conference on Genetic and
  evolutionary computation}, pages 845--852, 2011.

\bibitem[Schulman et~al.(2015)Schulman, Moritz, Levine, Jordan, and
  Abbeel]{schulman2015high}
John Schulman, Philipp Moritz, Sergey Levine, Michael Jordan, and Pieter
  Abbeel.
\newblock High-dimensional continuous control using generalized advantage
  estimation.
\newblock \emph{arXiv preprint arXiv:1506.02438}, 2015.

\bibitem[Silver et~al.(2014)Silver, Lever, Heess, Degris, Wierstra, and
  Riedmiller]{silver2014deterministic}
David Silver, Guy Lever, Nicolas Heess, Thomas Degris, Daan Wierstra, and
  Martin Riedmiller.
\newblock Deterministic policy gradient algorithms.
\newblock 2014.

\bibitem[Silver et~al.(2017)Silver, Schrittwieser, Simonyan, Antonoglou, Huang,
  Guez, Hubert, Baker, Lai, Bolton, et~al.]{silver2017mastering}
David Silver, Julian Schrittwieser, Karen Simonyan, Ioannis Antonoglou, Aja
  Huang, Arthur Guez, Thomas Hubert, Lucas Baker, Matthew Lai, Adrian Bolton,
  et~al.
\newblock Mastering the game of go without human knowledge.
\newblock \emph{Nature}, 550\penalty0 (7676):\penalty0 354--359, 2017.

\bibitem[Sutton and Barto(1998)]{sutton1998introduction}
Richard~S Sutton and Andrew~G Barto.
\newblock \emph{Introduction to Reinforcement Learning}, volume 135.
\newblock MIT press, 1998.

\bibitem[Sutton and Barto(2018)]{sutton2018reinforcement}
Richard~S Sutton and Andrew~G Barto.
\newblock \emph{Reinforcement Learning: An Introduction}.
\newblock MIT press, 2018.

\bibitem[Wei et~al.(2019)Wei, Zheng, Gayah, and Li]{wei2019survey}
Hua Wei, Guanjie Zheng, Vikash Gayah, and Zhenhui Li.
\newblock A survey on traffic signal control methods.
\newblock \emph{arXiv preprint arXiv:1904.08117}, 2019.

\bibitem[Wierstra et~al.(2014)Wierstra, Schaul, Glasmachers, Sun, Peters, and
  Schmidhuber]{wierstra2014natural}
Daan Wierstra, Tom Schaul, Tobias Glasmachers, Yi~Sun, Jan Peters, and
  J{\"u}rgen Schmidhuber.
\newblock Natural evolution strategies.
\newblock \emph{The Journal of Machine Learning Research}, 15\penalty0
  (1):\penalty0 949--980, 2014.

\bibitem[Williams(1992)]{williams1992simple}
Ronald~J Williams.
\newblock Simple statistical gradient-following algorithms for connectionist
  reinforcement learning.
\newblock \emph{Machine learning}, 8\penalty0 (3-4):\penalty0 229--256, 1992.

\end{thebibliography}
\bibliographystyle{plainnat}

\end{document}